# Shape and Centroid Independent Clustring Algorithm for Crowd Management Applications

Yasser Mohammad Seddiq, A. A. Alharbiy and Moayyad Hamza Ghunaim
King Abdulaziz City for Science and Technology (KACST)
Riyadh, Saudi Arabia
yseddiq@kacst.edu.sa, aharbiy@kacst.edu.sa, mghunaim@kacst.edu.sa

*Abstract*—Clustering techniques play an important role in data mining and its related applications. Among the challenging applications that require robust and real-time processing are crowd management and group trajectory applications. In this paper, a robust and low-complexity clustering algorithm is proposed. It is capable of processing data in a manner that is shape and centroid independent. The algorithm is of low complexity due to the novel technique to compute the matrix power. The algorithm was tested on real and synthetic data and test results are reported.

## I. Introduction

Clustering is the process of grouping similar objects together in an unsupervised way. Clustering algorithms play vital roles in data mining applications including group trajectory and crowd management. Those applications feature some unique challenges include unpredictable shapes of groups and real-time processing requirement. In crowd management, things change rapidly and response has to be very fast to avoid any catastrophic consequences in case of crowd crush. For those reasons, a clustering algorithms developed for crowd management application showed neutralize the effect of shape and it should feature low computational complexity.

Clustering is an *NP*-hard problem [1]. This has motivated the development of simplified algorithms that sacrifice performance to save time. Clustering algorithms can be grouped into different categories, the main two groups of them are:

1. Partitional clustering: where data are hardly partitioned into non-overlapped clusters, and usually the number of clusters is predefined. This group includes k-means family [2,3], provide simple and fast clustering, but it suffers from its random seeding techniques

2. Hierarchical clustering: which perform a multilevel clusteing either additively (bottom-up) by merging the nearest clusters; or divisively (top-down) by breaking down large cluster that contains sparse data. Although methods belong to this group achieve near-optimum results, it requires distance computation between every pair of samples, which is unreasonable in the case of large data [4].

The clustering is a function of similarity, cluster shape and centroid selection. [5]. The effect of cluster shape is very clear in the shape of data that forms a ring shape, in which the cluster centroid will exist in the middle where no data actually exist. While the selection of cluster centroid should guarantee that every group of data has a one and only one head, centroid selection methods (e.g. random generation, lowest ID or Nodal degree based) will never guarantee that as a centroid might be generated away from data or two centroid might be selected in the same cluster.

In this paper we present a method that can cancel the effect of the cluster head and cluster shape. Our method guarantees that a spatially close data of any shape will be grouped in one and only one cluster, and that cluster will include nothing but those data. The paper also involves proposing a novel method to calculate the matrix power which is essential to find the $k^{\text{th}}$ connection of a graph. Then that novel technique is used in the proposed clustering algorithm.

## II. The Proposed Clustring Algorithm

A clustering algorithm is proposed. It returns a class label for each one of the N input nodes of the set. It also ranks clusters upon their node count. This outcome of cluster size ranking is driven by the requirement of crowd management problem, where recognizing dense and big groups is of great interest since they are usually associated with accidents and crowd crush.

The algorithm parameter is the radius distance *r* that limits the size of the region of searching about neighbors. The algorithm proposed in this paper searches only within radius *r* and returns the found neighbors if any. The steps of the algorithm are explained in the following

### A. Adjacency Matrix Calculation

The algorithm starts by calculating the adjacency matrix $A$ of the $N$ nodes. $A$ is an $N{\times}N$ binary matrix where an element $A_{i,j} = 1$ indicates that the two nodes $p_i$ and $p_j$ are within distance *r* from one another. The Euclidian distance is used in calculating $A$.

## B. A Low-Complexity Novel Method for Matrix Power Calculation

From the adjacency matrix, the matrix power $G$ of size $N \times N$ is calculated. An ordinary way to calculate $G$ is

$$G = A^k, \quad \text{where } k = \lfloor N/2 \rfloor \quad (1)$$

This step is important in order to find the $k^{th}$ connection of graph at a certain node $p_i$ when finding all of its direct and indirect neighbors. That is, after rising to the power $k$, the algorithm will be able, in the upcoming steps, to determine the direct neighbor of a node $p_i$ (1st neighbor), the direct neighbors of the 1st neighbor (2nd neighbor), the direct neighbor of those 2st neighbor (3rd neighbor), and so on until the $k^{th}$ neighbor. The worst case is when the $N$ nodes are in a form of a chain such that each node $p_i$ has only two neighbors $p_{i-1}$ and $p_{i+1}$ where those two neighbor nodes are not neighbors of each other as illustrated in Fig. 1. The dashed lines in the figure represent indirect neighborhood when $k=1$. In such case, it is enough for the first and the last nodes to have at least one common indirect neighbor. In this example, both $p_1$ and $p_7$ have $p_4$ their 3rd indirect neighbor ($k = 3$). Hence, the algorithm will be able to place $p_1$ and $p_7$ in the same cluster. After calculating $G$, all elements of $G$ that are nonzero are set to 1 causing $G$ to be a binary matrix.

Calculating $G$ using (1) implies computational complexity of $O(N^N)$, which is a high complexity. Here, a novel less complex method of finding $G$ is proposed. Instead realizing (1) by means of repeatedly multiplying $A$ by itself $k$ times, a shortcut is applied by utilizing partial (intermediate) products. The first partial product that is produced is $A^2$. Instead of proceeding to multiply $A$ by $A^2$ to get $A^3$, we would utilize $A^2$ and multiply it by itself producing $A^4$. Likewise, $A^8$ is produced immediately from $A^4$. The method proceeds in calculating the square of the last partial product until the smallest product greater than $A^k$ is reached. That implies a number repeating multiplication $\lceil \log_2(\lfloor N/2 \rfloor) \rceil$ times where. Hence the complexity of this method is $O(N^m)$ where $m = \lceil \log_2(\lfloor N/2 \rfloor) \rceil$. A Comparison between the complexity of the proposed method and the ordinary one mentioned earlier is depicted in Fig. 2. The proposed method can be realized by the pseudocode in Fig. 3.

It is important to note that the results of (1) and the new method are not necessarily identical where the new method could produce results that are equal to greater than (1)'s. in fact, that acceptable in clustering applications. Recall that the value of $k$ in (1) is the minimum exponent to guarantee and common indirect neighbor between the first and the last nodes in a worst case scenario of a chain (Fig. 1). Raising $A$ to a power greater than $k$ yet satisfies the above condition of common neighbor and leads to the same algorithm outcomes without introducing any errors. However, in (1), increasing the value of $k$ increases the algorithm complexity further. On the other hand, the new method would produce results that (1) could produce with higher value of k, but with much less number of multiplications.

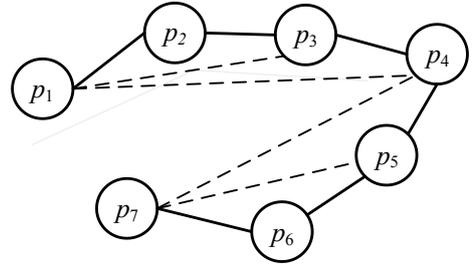

Fig. 1: Example of a node chain $k^{th}$ connection for $k=3$.

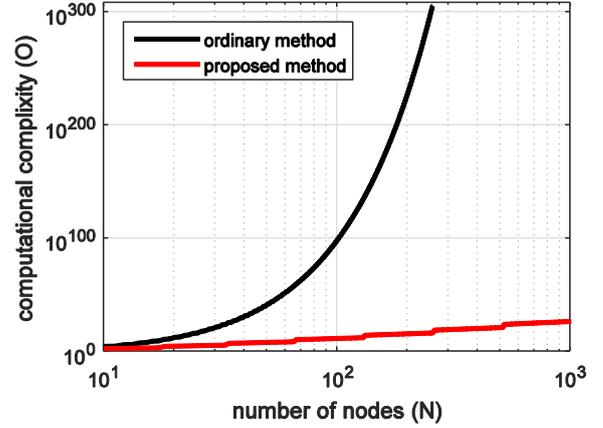

Fig. 2: Complexity of calculating matrix power

1. $m = \lceil \log_2(\lfloor N/2 \rfloor) \rceil$
2. $G = A$   // initialize G
3. for $i$=1:$m$
4. $\quad G = G^2$
5. end for

Fig. 3: Pseudocode of the proposed method to calculate matrix power

## C. Clustering Step

After the power matrix $G$ is calculated, the label vector $L$ is initialized to 0s. That is a vector of length $N$ where each element $l_i$ will be later assigned an integer-valued label to which the node $p_i$ belongs. Also, cluster counter $c$ is is initialized to 1. This is the integer that will be assigned to $l_i$. Now, the algorithm is ready to start clustering the nodes as follows:

1. For a node $p_i$, check if it has not been assigned to a cluster yet, i.e., its corresponding $l_i$ is equal to 0. If that is true, then set $l_i = c$ and proceed to 2. Otherwise, go to 4.
2. For $p_i$, the $i^{th}$ row of $G$ is used as mask vector $M$.
3. For all nodes $p_j$ that have not been assigned to a cluster yet, i.e., their corresponding $l_j$ is equal to 0, the vector $R$, which is the $j^{th}$ row of $G$, is masked by $M$. If the result is a zero vector, then $p_i$ and $p_j$ do not fall in the same cluster $c$. Otherwise, they belong to the same cluster and, hence, set $l_j = c$. Do this step for all $p_j$ whose label $l_j = 0$.
4. Increment $i$ and repeat steps 1 to 4 as long as there is at least one element in $L$ still equal to 0.

Finally, after all nodes are assigned to clusters, frequency of occurrence of each cluster label value is calculated and assigned to vector $F$. That is, each element $f_c$ of $F$ holds an integer showing how many nodes are in cluster $c$. From there can keep track of the largest cluster.

The pseudocode of the proposed algorithm is listed in Fig. 4. The bottleneck of this algorithm is the step of calculating the matrix power. Hence, the algorithm has complexity of $O(N^m)$ where $m = \lceil \log_2(\lfloor N/2 \rfloor) \rceil$.

### III. TESTING AND RESULTS

The algorithm was tested on real data and on synthetic data. Some synthetic data were generated to mimic some real-life scenarios of crowd behavior. On the other hand, the real data is GPS a set of coordinates of cars moving in a motorcade made for the purpose of this study. Clustering results presented in this section are directed graphically and color-coded upon cluster size. The largest cluster is marked in red. The next largest one is marked in green. The third largest is in blue. Other colors are used for smaller clusters, but we will focus on the three largest clusters.

Among the synthetic data, a hand-crafted data set of groups of points of various shapes was generated as depicted in Fig. 5. In Fig. 5 (a), a number of 7 clusters are returned. The largest cluster (red) is more like a crowd of 19 nodes. The second largest cluster (green) has 14 nodes in the worst case form discussed earlier, which is chain of single nodes. Cluster 3 (blue) has 13 nodes forming a shape of thicker chain than cluster 2.

The data of Fig. 5 (b) consists of 4 cluster. The largest was generated to span long distance and to contain branches that fork and then meet again. The remaining clusters are chains of various lengths.

Fig. 5 (c) illustrates a very important scenario in crowd management, which is highly dense crowd surrounded by lightly scattered nodes. That could represent many possible scenarios in real-life such as a group of people rushing to an exit, a mas fight, pedestrians who got their way blocked unexpectedly, …etc. it is clear how the algorithm managed to spot the danger zone which forms the largest cluster marked in red.

```
1.  for i = 1 : N
2.    for j = 1 : N
3.      d = √(Σ_{k=1}^{m} (p_{i,k} - p_{j,k})^2)   // Euclidian distance
4.      if d < r
5.        G_{i,j} = 1
6.      else
7.        G_{i,j} = 0
8.      end if
9.    end for
10. end for
          // calculate matrix power:
11. m = ⌈log_2(⌊N/2⌋)⌉
12. G = A     // initialize G
13. for i=1:m
14.   G = G^2
15. end for
16. Set all nonzero elements of G to 1 (0 otherwise).
    //end of calculating matrix power
17. for i = 1 : N
18.   L_i = 0   // initialize labels to 0
19. end for
20. i = 1   // initialize reference row counter
21. c = 1   // initialize cluster label number
22. while at least one label in L still equal to 0
23.   if L_i==0  //if this node hasn't been labeled yet
24.     M = the i^{th} row of G
25.     L_i = c   //use a new label number
26.     c = c + 1
27.     for j = i +1 : N
28.       if L_j==0
29.         R = the j^{th} row of G
30.         z = M AND R    // M ∩ R (mask R by M)
31.         if z ≠ 0  // not a zero vector ( M ∩ R ≠ Φ )
32.           L_j = L_i   // label node j with cluster# L_i
33.         end if
34.       end if
35.     end for
36.   end if
37.   i = i +1
38. end while
39. calculate F; the set of frequency of occurrence of each label
40. return L and F
```

Fig. 4: The proposed clustering algorithm

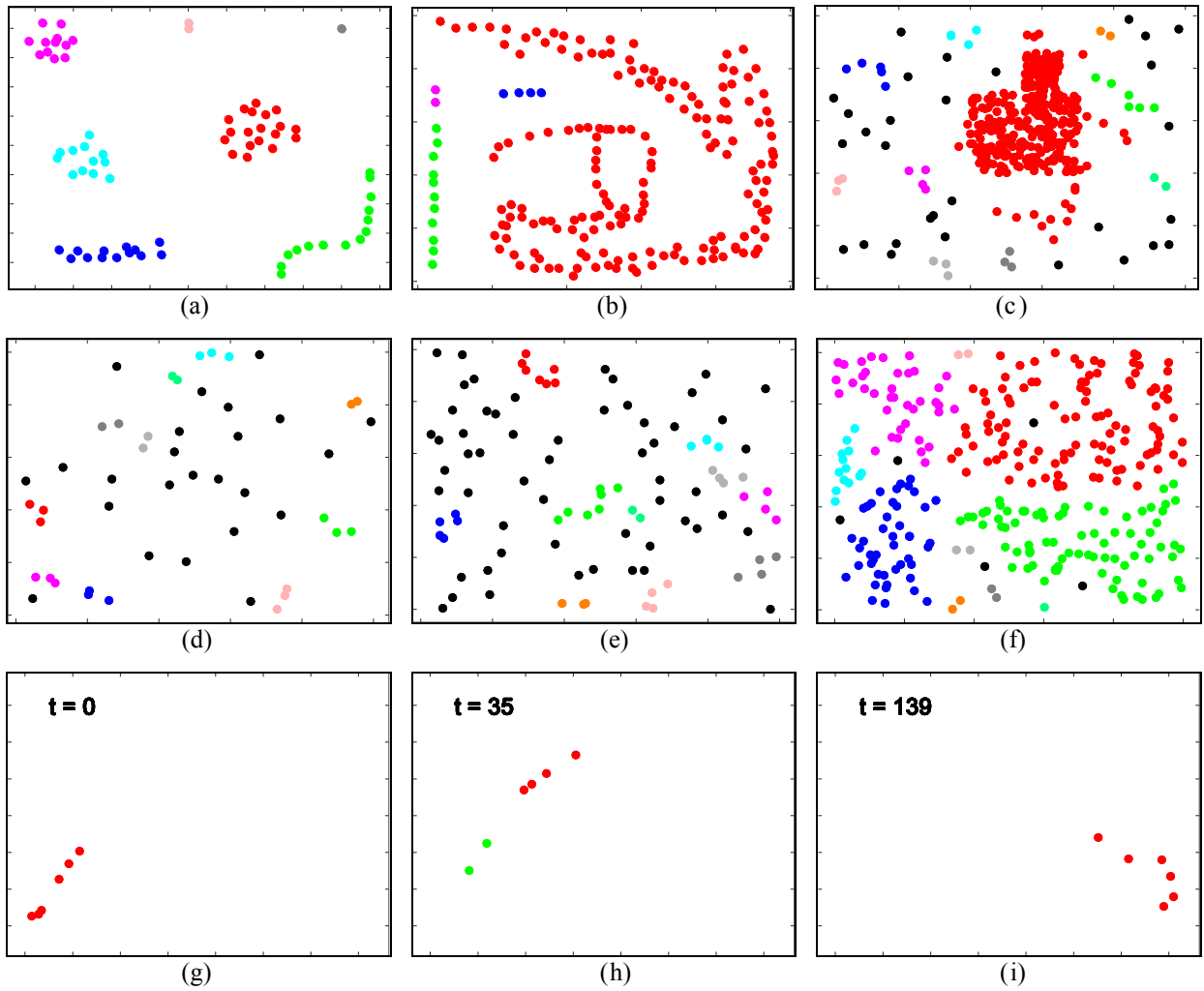

Fig. 5: Clustering algorithm test results

Fig. 5 (d)–(f) illustrates the algorithm outcomes of randomly scattered nodes of low, medium and high density respectively. In the low density crowd, there are many clusters of small number (around 3 nodes per cluster for the first few largest clusters). When the density increases, that ratio becomes around 7 per cluster. For the high density data, the number clusters is less despite the high total node count. That is because nodes are within very short distance from each other.

Samples of the real motorcade GPS data are illustrated in Fig. 5 (g)–(i). Selected time instances showing when some nodes join and disjoin the group. At the initial time instance $t_0$, all cars started form rest. At instance $t_{35}$, two cars have been in a distance that is long enough form others to be recognized by the algorithm as a separate cluster. Later at $t_{139}$, they rejoined again forming one cluster. An animated illustration of full trip clustering can be downloaded at https://www.dropbox.com/s/min70nkvkzgem9x/fullTripClustring.gif?dl=0.

ACKNOWLEDGMENT

Grateful acknowledgment to the volunteers who assisted in collecting the test data for this work.